\documentclass{article}
\usepackage{ijcai26}

\usepackage{times}
\usepackage{soul}
\usepackage{url}
\usepackage[hidelinks]{hyperref}
\usepackage[utf8]{inputenc}
\usepackage[small]{caption}
\usepackage{graphicx}
\usepackage{amsmath}
\usepackage{amsthm}
\usepackage{booktabs}
\usepackage{algorithm}
\usepackage{algorithmic}
\usepackage[switch]{lineno}
\usepackage{xcolor}
\usepackage{tikz}

\newlength{\SRcellw}
\newlength{\CTcellw}
\newlength{\TPcellw}
\newcommand{\heatcell}[5]{%
  \pgfmathsetmacro{\pct}{100*(#1-#2)/(#3-#2)}%
  \pgfmathsetmacro{\pct}{max(0,min(100,\pct))}%
  \ifnum#4=0\relax
    \pgfmathsetmacro{\pct}{100-\pct}%
  \fi
  \edef\cellcolor{green!\pct!red}%
  \tikz[baseline=(X.base)]\node[fill=\cellcolor, fill opacity=0.25, text opacity=1,
    rounded corners=1.5pt, inner xsep=2.2pt, inner ysep=0.8pt,
    minimum width=#5, text width=#5, align=center](X){\strut #1};%
}
\newcommand{\SRcell}[1]{\heatcell{#1}{72.9}{100.0}{1}{\SRcellw}}
\newcommand{\CTcell}[1]{\heatcell{#1}{0.19}{14.15}{0}{\CTcellw}}
\newcommand{\TPcell}[1]{\heatcell{#1}{0.10}{0.45}{1}{\TPcellw}}

\title{It Takes Two to Tango: A Holistic Simulator for Joint Order Scheduling and Multi-Agent Path Finding in Robotic Warehouses}

\author{%
Haozheng Xu$^{1}$ \and Wenhao Li$^{2}$ \and Zifan Wei$^{1}$ \and Bo Jin$^{2}$\\
Hongxing Bai$^{3}$ \and Ben Yang$^{1}$ \and Xiangfeng Wang$^{1,4}$\\
\affiliations
$^{1}$East China Normal University \quad
$^{2}$Tongji University \quad \\
$^{3}$Zhejiang Galaxis Technology Group Co.,Ltd \quad
$^{4}$Shenzhen Loop Area Institute (SLAI)\\
}

\begin{document}

\maketitle

\begin{abstract}
The prevailing paradigm in Robotic Mobile Fulfillment Systems (RMFS) typically treats order scheduling and multi-agent pathfinding as isolated sub-problems.
We argue that this decoupling is a fundamental bottleneck, masking the critical dependencies between high-level dispatching and low-level congestion.
Existing simulators fail to bridge this gap, often abstracting away heterogeneous kinematics and stochastic execution failures.
We propose {\bf{WareRover}}, a holistic simulation platform that enforces a tight coupling between OS and MAPF via a unified, closed-loop optimization interface.
Unlike standard benchmarks, WareRover integrates dynamic order streams, physics-aware motion constraints, and non-nominal recovery mechanisms into a single evaluation loop.
Experiments reveal that SOTA algorithms often falter under these realistic coupled constraints, demonstrating that WareRover provides a necessary and challenging testbed for robust, next-generation warehouse coordination.
The project and video is available at \url{https://hhh-x.github.io/WareRover/}.

\end{abstract}

\section{Introduction}

The exponential surge in e-commerce has necessitated a paradigm shift in warehouse logistics, driving the transition from manual operations to high-density automation~\cite{da2021robotic,lamballais2017estimating,zhen2023deploy}.
With order picking constituting $50\%$--$65\%$ of operational costs~\cite{xie2021introducing}, RMFS have emerged as the industry standard.
By employing a goods-to-person workflow where AGVs transport shelves to picking stations~\cite{yuan2017bot}, RMFS achieves 2--3$\times$ the productivity of manual approaches, a capability leveraged by industry giants like Amazon, JD.com, and Cainiao~\cite{merschformann2018multi,cheng2024deep}.

Central to RMFS efficiency is Multi-Agent Pathfinding (MAPF), which governs the coordination of massive AGVs~\cite{chen2024traffic}.
However, real-world deployment introduces complexities often abstracted away in theoretical models: fleets are increasingly heterogeneous—handling diverse tasks from pod transport to pallet replenishment—imposing distinct kinematic and operational constraints~\cite{chen2026multi,lin2024multi}.
Furthermore, the environment is fundamentally dynamic; AGV failures, communication latencies, and traffic congestion necessitate robust replanning in continuous time~\cite{bertoli2024fault}.
Consequently, optimal performance requires a holistic system where collision-free planning, order scheduling, and execution feedback are not treated in isolation, but as a tightly coupled, end-to-end loop.

Despite this necessity, a critical dichotomy exists in current simulation methodologies.
On one hand, RMFS-centric platforms like RAWSim-O~\cite{merschformann2017rawsim} excel in high-level order and inventory modeling~\cite{xie2021introducing,silva2024manual} but often trivialize pathfinding via heuristic or rule-based strategies~\cite{merschformann2019decision}.
Crucially, these systems exhibit a weak coupling between task scheduling and agent navigation, masking the impact of congestion on schedule feasibility.
On the other hand, MAPF-centric simulators such as Flatland~\cite{mohanty2020flatland} and POGEMA~\cite{skrynnik2024pogema} prioritize algorithmic evaluation but operate in abstract, homogeneous grid worlds.
Even recent platforms like SkyRover~\cite{ma2025skyrover}, which introduce heterogeneity, still abstract away the semantic complexities of warehouse logic, such as multi-stage workflows and dynamic order injection.
To the best of our knowledge, no existing platform bridges this gap to support advanced MAPF evaluation within a semantically realistic, operationally constrained RMFS environment.


This disconnect is non-trivial. 
Warehouse MAPF is substantially harder than generic grid-based pathfinding due to the interplay of dense layouts, heterogeneous kinematics, and physical execution limits (e.g., turning costs, velocity bounds).
Moreover, the assumption of nominal execution in standard algorithms~\cite{wang2025paths,sharon2015conflict,chan2023greedy} collapses under real-world stochasticity, where mechanical failures or delays invalidate static plans~\cite{andreychuk2022multi}.
The absence of a unified simulation testbed thus hinders the development of algorithms capable of surviving the ``sim-to-real'' transfer.

To bridge this divide, we present {\bf{WareRover}}, a holistic RMFS simulation platform designed to facilitate joint research on order scheduling and robust pathfinding under both nominal and non-nominal conditions.
By enforcing a tight integration between high-level logistics logic and low-level motion planning, WareRover enables the evaluation of the full operational stack.
Our key contributions include:
1) {\bf{Closed-Loop Joint Optimization}}: A unified interface coupling order scheduling with MAPF, enabling the study of scheduling strategies that explicitly account for traffic and routing costs;
2) {\bf{High-Fidelity Warehouse Modeling}}: A topology-agnostic framework supporting customizable layouts, heterogeneous fleets, and realistic order streams;
3) {\bf{Resilience Evaluation}}: Native support for simulating stochastic failures (e.g., breakdowns, delays), allowing researchers to benchmark algorithmic robustness against execution uncertainty.

\section{Related Work}

\noindent {\bf{RMFS and MAPF Simulators}}: Existing warehouse robotics simulators generally fall into two categories: RMFS-oriented platforms and MAPF-centric simulators.The RMFS-oriented platforms, such as RAWSim-O~\shortcite{merschformann2017rawsim}, Alphabet Soup~\cite{hazard2006alphabet}, and industrial tools~\cite{benavides2024robotic}, focus on order generation and end-to-end fulfillment workflows. 
However, they typically decouple order scheduling and path finding, providing limited support for their joint optimization. 
In addition, their path finding modules are often simplified, with limited support for large-scale conflict-free MAPF, heterogeneous AGVs, or execution-level failures.
The MAPF-centric simulators, including Flatland~\shortcite{mohanty2020flatland}, POGEMA~\shortcite{skrynnik2024pogema}, mapf-IR~\cite{okumura2021iterative}, SMART~\cite{yan2025advancing}, and SkyRover~\cite{ma2025skyrover}, emphasize algorithm-centric benchmarking in abstract grid-based environments with detailed modeling of agent interactions. 
Nevertheless, they largely lack warehouse-specific semantics, such as order scheduling, multi-stage tasks, and structured storage layouts~\cite{sun2025benchmark}, limiting their applicability to realistic RMFS settings.

\smallskip
\noindent {\bf{MAPF Algorithms in Warehouse Settings}}: MAPF in warehouse environments has received increasing attention~\cite{wang2024mapf}. 
Most studies build on classical search-based methods, such as A* and CBS~\shortcite{sharon2015conflict}, or incorporate heuristic and learning-based strategies to improve performance in dense settings~\cite{lim2025cbs,veerapaneni2023effective,ma2021distributed}. 
However, these approaches are typically evaluated under simplified assumptions, with limited modeling of warehouse-specific factors such as dynamic order arrivals, heterogeneous AGVs, physical execution constraints, and execution failures. 
As a result, the applicability of these methods in real RMFS deployments remains insufficiently validated.
In contrast, WareRover is designed specifically for RMFS scenarios and integrates warehouse-specific task modeling with execution-aware multi-agent path finding under both nominal and non-nominal conditions. 
It supports heterogeneous AGVs, joint evaluation of order scheduling and path finding, online replanning, and explicit modeling of AGV failure and recovery, providing a unified platform for evaluating MAPF methods in realistic warehouse settings.

\section{The WareRover Platform}
\begin{figure}[t]
    \centering
    \includegraphics[width=\linewidth]{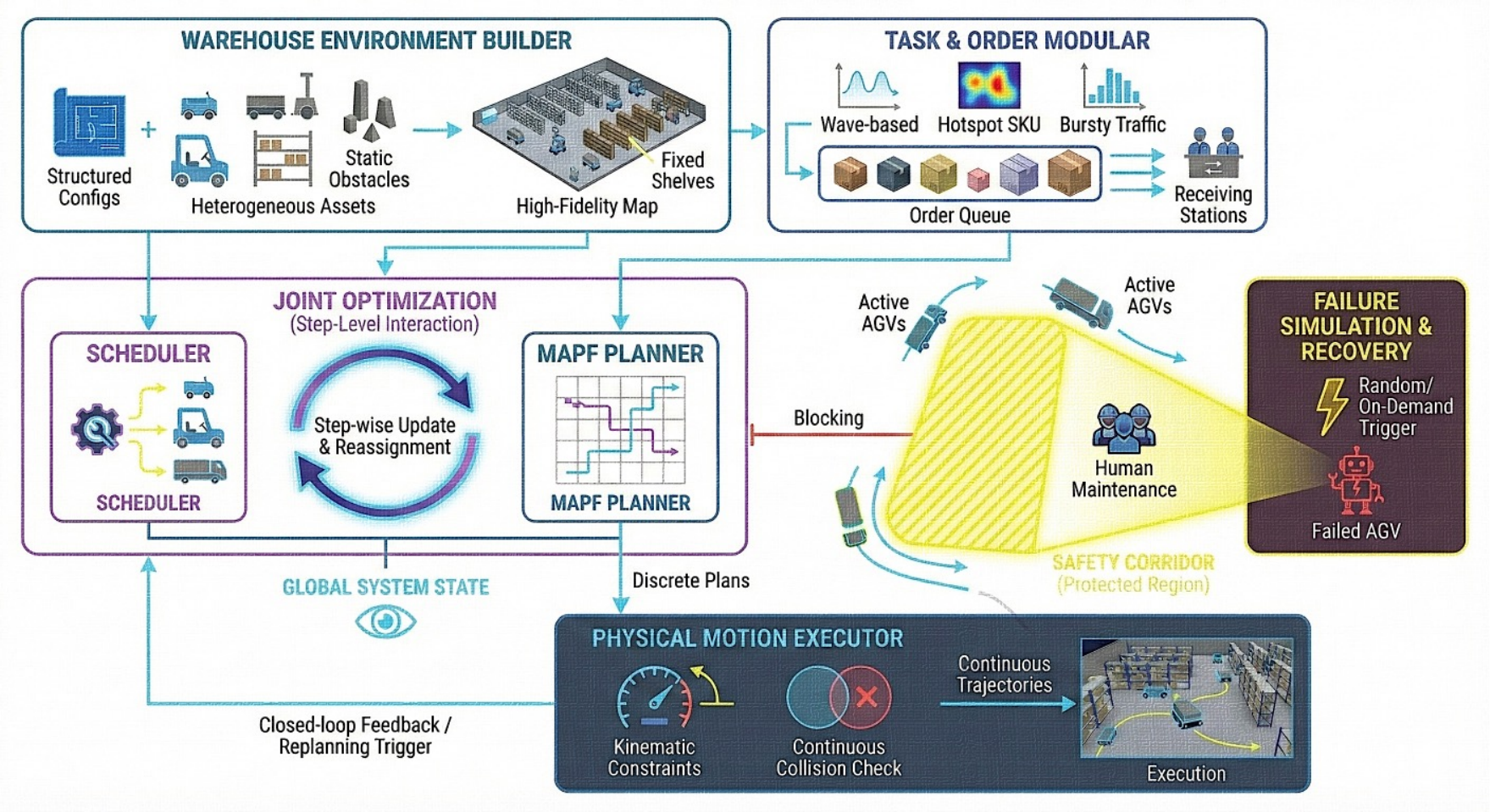}
    \caption{The WareRover integrates (i) a warehouse environment builder for structured configurations and high-fidelity maps, (ii) a task \& order module for realistic order streams and task decomposition, (iii) a step-level joint optimization loop that couples the scheduler and MAPF planner via global state feedback and online replanning, (iv) a physical motion executor that enforces kinematic constraints with continuous collision checking, and (v) failure simulation \& recovery that triggers maintenance corridors and rerouting under non-nominal execution.}
    \label{fig:system_overview}
\end{figure}

Figure~\ref{fig:system_overview} provides an overview of the WareRover platform, illustrating both the heterogeneous warehouse environment and the built-in failure simulation and recovery mechanism.

\begin{figure*}[t]
    \centering
    \includegraphics[width=\textwidth]
    {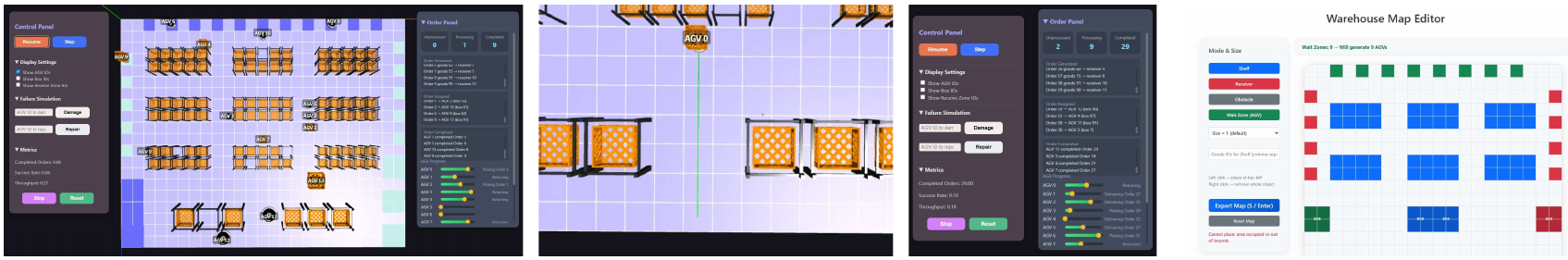}
    \caption{System demonstration of WareRover.
    From left to right: (a) real-time warehouse simulation with AGV motion and task execution;
    (b) failure injection and safety corridor generation under AGV breakdown;
    (c) interactive control panel for experiment management and parameter configuration;
    (d) warehouse environment editor for structured map construction and layout configuration.}
    \label{fig:system_demo}
\end{figure*}

\smallskip
Figure~\ref{fig:system_demo} complements the architecture by presenting the integrated system visualization and experimental workflow of WareRover, including visualization, control, failure simulation, and environment editing.

\noindent {\bf{Warehouse Environment Builder}}: To bridge the gap between abstract MAPF benchmarks and real-world RMFS scenarios, WareRover provides a configurable warehouse environment builder. 
Warehouse layouts are specified through structured configurations, including aisles, shelves, receiving stations, and AGV parking areas, enabling the modeling of realistic warehouse facilities.
Unlike grid-based simulators, WareRover supports high-fidelity environments composed of static elements such as shelves and fixed obstacles, as illustrated in Figure~\ref{fig:system_overview}(a).
The platform further supports heterogeneous AGVs and cargo units by explicitly modeling physical size, velocity, and operational constraints, and allows layouts to scale from small test maps to industrial-scale warehouses.

\smallskip
\noindent {\bf{Task and Order Modular}}: WareRover integrates an order- and task-management modular to model order-driven logistics in realistic warehouse environments. 
The system supports multi-stage operations such as picking and delivery, and can simulate representative e-commerce workloads, including wave-based orders, hotspot SKUs with skewed demand, and bursty promotional traffic.
Each order contains structured attributes such as item type, container size, and target receiving station. 
Since containers may store multiple items, task scheduling involves joint decisions over AGV assignment and item selection, resulting in a tightly coupled interaction between task decomposition and path planning.
The scheduler dynamically maintains task queues based on AGV states and exposes extensible interfaces for custom scheduling.

\smallskip
\noindent {\bf{Joint Optimization}}: WareRover adopts a unified joint optimization procedure that tightly couples task scheduling and multi-agent path finding through step-level interaction.
At each simulation step, the scheduler is invoked to update AGV task queues based on the global system state, enabling flexible task reassignment over time.
Given the current assignments, the path planner generates collision-free trajectories and performs online replanning when congestion, blockage, or idle states occur.
Both the scheduler and planner can access global environment and agent information, facilitating future planning and scheduling strategies that explicitly reason about system-wide interactions.
Although the algorithms implemented in this work focus on idle AGVs and local replanning, the modular yet tightly coupled interface design provides a flexible foundation for more advanced joint optimization and learning-based approaches.

\smallskip
\noindent {\bf{Physical Motion Executor}}: The physical motion executor converts discrete plans into continuous trajectories that satisfy kinematic constraints.
It performs continuous-time collision checking among AGVs, shelves, and obstacles, enforcing velocity limits, turning constraints, and safety margins.
When infeasible execution or severe congestion is detected, the executor triggers replanning, forming closed-loop feedback between discrete planning and continuous execution.
This design enables the evaluation of MAPF algorithms under realistic motion constraints beyond idealized settings.

\smallskip
\noindent {\bf{Failure Simulation}}: WareRover includes a failure simulation and safety recovery module to evaluate the robustness of scheduling and MAPF under abnormal conditions.
AGV failures can be triggered randomly or on demand. 
Upon failure, the system generates a protected safety corridor for human maintenance, while the affected AGV is treated as unavailable, as illustrated in Fig.~\ref{fig:system_overview}(b).
The safety corridor is respected as a region in subsequent path finding, requiring other AGVs to actively avoid it.
By modeling AGV failures, temporary resource unavailability, and localized blockages, WareRover enables evaluation of recovery behavior and system stability under realistic operating conditions.

\section{Experiments}
We evaluate the WareRover following:
1) $3$ scenarios: homogeneous, heterogeneous, and fault-tolerant;
2) $2$ order scheduling strategies: order scheduling (TA) and random dispatching (RD);
3) $3$ MAPF algorithms: A*, Conflict-Based Search (CBS), and DHC;
4) $5$ order generation patterns with each configuration repeated 100 times.
Performance is measured using success rate (SR), computation time (CT), and throughput (TP).
Complete results and additional visualizations are available on our website and video documentation.

\smallskip
\noindent {\bf{Homogeneous Environment}}: All AGVs and cargo share identical physical dimensions.
The map size is $20\times15$ with $9$ AGVs, $32$ shelves, and $8$ receiving areas, under one-shot order generation (OS).
In Table~\ref{tab:env_results} (Homogeneous), all method combinations achieve near-perfect success rates in this simple setting.
TA combined with CBS achieves the highest throughput with minimal computation time serving as a baseline. 

\begin{table}[htb!]
    \centering
    \resizebox{.8\linewidth}{!}{%
    {\renewcommand{\arraystretch}{1.15}%
    \begin{tabular}{llrrr}
        \toprule
        Env. & Method & SR (\%) & CT & TP \\
        \midrule
        Ho & RD + A*  & \SRcell{100.0} & \CTcell{0.53}  & \TPcell{0.21} \\
        Ho & RD + CBS & \SRcell{100.0} & \CTcell{0.55}  & \TPcell{0.22} \\
        Ho & RD + DHC & \SRcell{99.9}  & \CTcell{7.70}  & \TPcell{0.21} \\
        Ho & TA + A*  & \SRcell{100.0} & \CTcell{0.34}  & \TPcell{0.34} \\
        Ho & TA + CBS & \SRcell{100.0} & \CTcell{0.31}  & \TPcell{0.36} \\
        Ho & TA + DHC & \SRcell{100.0} & \CTcell{4.87}  & \TPcell{0.33} \\
        He & TA + A*  & \SRcell{100.0} & \CTcell{0.19}  & \TPcell{0.42} \\
        He & TA + CBS & \SRcell{98.8}  & \CTcell{0.40}  & \TPcell{0.45} \\
        He & TA + DHC & \SRcell{72.9}  & \CTcell{9.52}  & \TPcell{0.10} \\
        FT & TA + A*  & \SRcell{100.0} & \CTcell{1.45}  & \TPcell{0.25} \\
        FT & TA + CBS & \SRcell{99.9}  & \CTcell{14.15} & \TPcell{0.26} \\
        FT & TA + DHC & \SRcell{98.9}  & \CTcell{7.93}  & \TPcell{0.20} \\
        \bottomrule
    \end{tabular}}%
    }
    \caption{Performance comparison across three environments under one-shot order generation (OS). Env.: Ho=Homogeneous, He=Heterogeneous, FT=Fault-tolerant.}
    \label{tab:env_results}
\end{table}

\smallskip
\noindent {\bf{Heterogeneous Environment}}: 
Multiple mixed AGVs exsit with two cargo sizes ($1\times 1$ and $2\times 2$), reflecting realistic RMFS deployments.
Table~\ref{tab:env_results} shows that TA combined with A* and CBS maintains high SR, while DHC suffers significant degradation, particularly for larger agents.
CBS exhibits a moderate success-rate drop due to increased spatial conflicts, but still substantially outperforms DHC, compared to homogeneous.
These indicate that explicit conflict reasoning becomes increasingly important in heterogeneous.

\smallskip
\noindent {\bf{Fault-Tolerant Environment}}: To evaluate robustness, we introduce temporary AGV failures during execution.
An active AGV fails with probability $1\%$, and remains inactive for $40$ steps, and then recovers, inducing frequent replanning.
The behaviors of algorithms differ markedly from static scenarios (Table~\ref{tab:env_results}).
The A* achieves the lowest computation time due to lightweight local replanning, whereas CBS incurs significantly higher cost under frequent failures.
DHC maintains moderate computation time with higher SR.
Replanning efficiency is a dominant factor in fault-prone environments.

\clearpage
\newpage

\bibliographystyle{named}
\bibliography{ijcai26}

@article{da2021robotic,
  title={Robotic mobile fulfillment systems: A survey on recent developments and research opportunities},
  author={da Costa Barros, {\'I}talo Renan and Nascimento, Tiago Pereira},
  journal={Robotics and Autonomous Systems},
  volume={137},
  pages={103729},
  year={2021}
}

@article{lamballais2017estimating,
  title={Estimating performance in a robotic mobile fulfillment system},
  author={Lamballais, Tim and Roy, Debjit and others},
  journal={European Journal of Operational Research},
  volume={256},
  number={3},
  pages={976--990},
  year={2017}
}

@article{zhen2023deploy,
  title={How to deploy robotic mobile fulfillment systems},
  author={Zhen, Lu and Tan, Zheyi and others},
  journal={Transportation Science},
  volume={57},
  number={6},
  pages={1671--1695},
  year={2023}
}

@article{xie2021introducing,
  title={Introducing split orders and optimizing operational policies in robotic mobile fulfillment systems},
  author={Xie, Lin and Thieme, Nils and others},
  journal={European Journal of Operational Research},
  volume={288},
  number={1},
  pages={80--97},
  year={2021}
}

@article{yuan2017bot,
  title={Bot-in-time delivery for robotic mobile fulfillment systems},
  author={Yuan, Zhe and Gong, Yeming Yale},
  journal={IEEE Transactions on Engineering Management},
  volume={64},
  number={1},
  pages={83--93},
  year={2017}
}

@article{merschformann2018multi,
  title={Multi-agent path finding with kinematic constraints for robotic mobile fulfillment systems},
  author={Merschformann, Marius and Xie, Lin and others},
  article={arxiv preprint arXiv:1706.09347},
  year={2018}
}

@article{cheng2024deep,
  title={Deep reinforcement learning driven cost minimization for batch order scheduling in robotic mobile fulfillment systems},
  author={Cheng, Bayi and Xie, Tao and others},
  journal={Expert Systems with Applications},
  volume={255},
  pages={124589},
  year={2024}
}

@inproceedings{chen2024traffic,
  title={Traffic flow optimisation for lifelong multi-agent path finding},
  author={Chen, Zhe and Harabor, Daniel and others},
  booktitle={AAAI},
  year={2024}
}

@article{merschformann2019decision,
  title={Decision rules for robotic mobile fulfillment systems},
  author={Merschformann, Marius and Lamballais, Tim and others},
  journal={Operations Research Perspectives},
  volume={6},
  pages={100128},
  year={2019}
}

@article{merschformann2017rawsim,
  title={RAWSim-O: A simulation framework for robotic mobile fulfillment systems},
  author={Merschformann, Marius and Xie, Lin and others},
  journal={arXiv preprint arXiv:1710.04726},
  year={2017}
}

@article{silva2024manual,
  title={Manual and robotic storage and picking systems: a literature review},
  author={Silva, Allyson and Coelho, Leandro C and others},
  journal={INFOR: Information Systems and Operational Research},
  volume={62},
  number={3},
  pages={313--343},
  year={2024}
}

@article{skrynnik2024pogema,
  title={{POGEMA}: A benchmark platform for cooperative multi-agent pathfinding},
  author={Skrynnik, Alexey and Andreychuk, Anton and others},
  journal={arXiv preprint arXiv:2407.14931},
  year={2024}
}

@article{mohanty2020flatland,
  title={Flatland-{RL}: Multi-agent reinforcement learning on trains},
  author={Mohanty, Sharada and Nygren, Erik and others},
  journal={arXiv preprint arXiv:2012.05893},
  year={2020}
}

@inproceedings{ma2025skyrover,
  title     = {SkyRover: A Modular Simulator for Cross-Domain Pathfinding},
  author    = {Ma, Wenhui and Li, Wenhao and others},
  booktitle = {IJCAI},
  year      = {2025}
}

@article{sharon2015conflict,
  title={Conflict-based search for optimal multi-agent pathfinding},
  author={Sharon, Guni and Stern, Roni and others},
  journal={Artificial intelligence},
  volume={219},
  pages={40--66},
  year={2015}
}

@inproceedings{chan2023greedy,
  title={Greedy priority-based search for suboptimal multi-agent path finding},
  author={Chan, Shao-Hung and Stern, Roni and others},
  booktitle={SoCS},
  year={2023}
}

@article{andreychuk2022multi,
  title={Multi-agent pathfinding with continuous time},
  author={Andreychuk, Anton and Yakovlev, Konstantin and others},
  journal={Artificial Intelligence},
  volume={305},
  pages={103662},
  year={2022}
}

@inproceedings{hazard2006alphabet,
  title={Alphabet {S}oup: A testbed for studying resource allocation in multivehicle systems},
  author={Hazard, Christopher J and Wurman, Peter R and others},
  booktitle={AAAI Workshop},
  year={2006}
}

@article{benavides2024robotic,
  title={Robotic mobile fulfillment system: a systematic review},
  author={Benavides-Robles, Maria Torcoroma and Valencia-Rivera, Gerardo Humberto and others},
  journal={IEEE Access},
  volume={12},
  pages={16767--16782},
  year={2024}
}

@inproceedings{okumura2021iterative,
  title={Iterative refinement for real-time multi-robot path planning},
  author={Okumura, Keisuke and Tamura, Yasumasa and others},
  booktitle={IROS},
  year={2021}
}

@article{yan2025advancing,
  title={Advancing {MAPF} towards the real world: A scalable multi-agent realistic testbed ({SMART})},
  author={Yan, Jingtian and Li, Zhifei and others},
  journal={arXiv preprint arXiv:2503.04798},
  year={2025}
}

@article{sun2025benchmark,
  title={Benchmark for multi-agent pickup and delivery problem in a robotic mobile fulfillment system},
  author={Sun, Yangjun and Zhao, Ning},
  journal={Flexible Services and Manufacturing Journal},
  volume={37},
  number={3},
  pages={697--729},
  year={2025}
}

@inproceedings{wang2024mapf,
  title={{MAPF} in {3D} warehouses: dataset and analysis},
  author={Wang, Qian and others},
  booktitle={ICAPS},
  year={2024}
}

@article{lim2025cbs,
  title={{CBS-Budget} ({CBSB}): A complete and bounded suboptimal search for multi-agent path finding},
  author={Lim, Jaein and Tsiotras, Panagiotis},
  journal={Artificial Intelligence},
  pages={104349},
  year={2025}
}

@inproceedings{veerapaneni2023effective,
  title={Effective integration of weighted cost-to-go and conflict heuristic within suboptimal {CBS}},
  author={Veerapaneni, Rishi and Kusnur, Tushar and others},
  booktitle={AAAI},
  year={2023}
}

@inproceedings{ma2021distributed,
  title={Distributed heuristic multi-agent path finding with communication},
  author={Ma, Ziyuan and Luo, Yudong and others},
  booktitle={ICRA},
  year={2021}
}

@article{chen2026multi,
  title={Multi-agent task assignment and path finding for heterogeneous AGV systems},
  author={Chen, Feifei and Li, Li and Tang, Junya},
  journal={Expert Systems with Applications},
  pages={130973},
  year={2026}
}

@article{lin2024multi,
  title={Multi-Agent Path Finding With Heterogeneous Geometric and Kinematic Constraints in Continuous Space},
  author={Lin, Wenbo and Song, Wei and others},
  journal={RAL},
  year={2025},
  volume={10},
  issue={1},
  pages={492--499}
}

@article{bertoli2024fault,
  title={Fault diagnosis and identification in {AGV}s system},
  author={Bertoli, Annalisa and Battilani, Nicola and others},
  journal={IFAC-PapersOnLine},
  volume={58},
  number={4},
  pages={246--251},
  year={2024}
}

@article{wang2025paths,
  title={Where Paths Collide: A Comprehensive Survey of Classic and Learning-Based Multi-Agent Pathfinding},
  author={Wang, Shiyue and Xu, Haozheng and others},
  journal={arXiv preprint arXiv:2505.19219},
  year={2025}
}

\end{document}